\documentclass[letterpaper, 10 pt, conference]{ieeeconf}  

\IEEEoverridecommandlockouts                              

\usepackage{graphics} 
\usepackage{epsfig} 
\usepackage{mathptmx} 
\usepackage{times} 
\usepackage{amsmath} 
\usepackage{amssymb}  
\usepackage{booktabs}
\usepackage{pifont}
\usepackage{multirow}
\usepackage{tabularx}

\title{\LARGE \bf
Robust Slip Detection and Material Classification via Spatiotemporal Transformers on a Uniformly-Illuminated Visuo-Tactile Sensor}

\author{Ziyang Ma, Yuhao Sun, Zichen Ai, Xiangyang Ji*, Bin Fang*
\thanks{This work was supported by Brain Science and Brain-like Intelligence Technology - National Science and Technology Major Project (Grant No. 2025ZD0215600),  in part by the National Natural Science Foundation of China under Grant No.62573063, 62536001 and the Open Foundation of the State Key Laboratory of Precision Space-time Information Sensing Technology No.STSL2025-B-07-01(C)
(Ziyang Ma, Yuhao Sun and Zichen Ai contributed equally to this work.)(Corresponding authors: Xiangyang Ji; Bin Fang.)}%
\thanks{Ziyang Ma, Yuhao Sun, Zichen Ai and Bin Fang are with the School of Artificial Intelligence, Beijing University of Posts and Telecommunications, Beijing 100876, China (e-mail: maziyang@bupt.edu.cn; yuhaosun@bupt.edu.cn; aizichen@bupt.edu.cn; fangbin1120@bupt.edu.cn).}
}

\begin{document}

\maketitle
\thispagestyle{empty}
\pagestyle{empty}

\begin{abstract}
Tactile sensing is central to robotic manipulation, among which slip detection stands out as a quintessential and critical task. However, existing slip datasets are predominantly limited to binary classification, lacking fine-grained directional perception. To address this limitation, we propose a visuo-tactile sensor featuring customized uniform RGB illumination, alongside a unified perception framework. At the hardware level, the sensor achieves high-precision, sub-millimeter depth reconstruction. Based on this capability, we collect a multi-task visuo-tactile dataset encompassing 15 objects, synchronously generating depth information for each data sample. Algorithmically, we design a dual-head TimeSformer network to process dynamic spatiotemporal slip. On unseen objects, this network achieves robust accuracies of 95.5\% and 91.5\% for 3-class contact state prediction and fine-grained 8-class slip direction classification, respectively. Furthermore, static tactile-based object class recognition utilizing a ResNet-50 backbone yields an outstanding accuracy of 98.8\% across 15 categories. The proposed hardware-software framework provides high-fidelity feedback and a powerful multi-modal perception baseline for complex robotic manipulation.
\end{abstract}

\section{INTRODUCTION}

Tactile sensing serves as a fundamental modality for robots to interact with the physical world, analogous to the human sense of touch \cite{dahiya2010tactile, li2020review}. %
Unlike computer vision, which acquires global geometric information from a distance, tactile sensing provides direct, high-fidelity feedback during the physical contact process \cite{luo2017robotic}.
It plays a crucial role in closed-loop manipulation tasks, enabling robots to perceive local surface textures, estimate contact forces, and ensure stable grasping \cite{calandra2018more,chen2023plasticine}.   
Among various tactile sensing technologies, visuo-tactile sensors (VTS) have emerged as a dominant research direction \cite{yuan2017gelsight, ward2018tactip,zhang2022hardware,Li_2025, sun2025tactile}.   
By employing high-resolution embedded cameras to monitor the deformation of soft elastomers, VTS convert tactile stimuli into rich visual data, offering a spatial resolution that far exceeds traditional electronic skins or piezoresistive arrays \cite{zhang2025artificialskin}.  

\begin{figure}[htbp]
    \centering
    \includegraphics[width=1\linewidth]{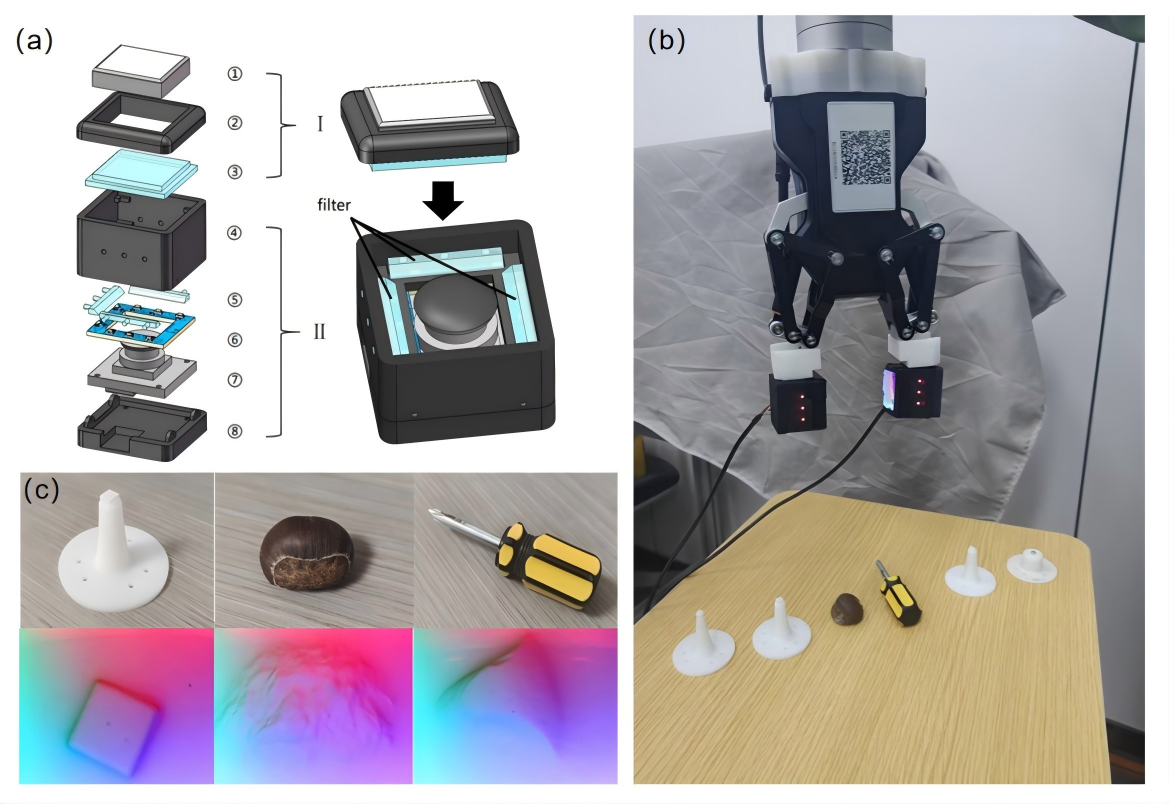} 
    \caption{(a) Overall structural diagram of the proposed sensor.(b) Dataset collection using the visuotactile sensor for slip status detection.(c) Actual tactile images collected by pressing 15 common objects. These images constitute a slip detection dataset of 7,925 samples and a high-precision material classification dataset of approximately 19,336 images. (The figure shows some of the collected objects.)} %
    \label{fig:sensor_overview}
\end{figure}

The inherent advantage of visuo-tactile sensors lies in their ability to leverage mature computer vision algorithms for multimodal perception \cite{lepora2021soft}. %
Specifically, slip detection and tactile-based Object Class recognition are two critical capabilities for intelligent manipulation. %
Real-time slip detection allows a robotic gripper to dynamically adjust its grasping force before an object drops \cite{zhang2018slip}, while tactile-based object class recognition enables the robot to infer the physical properties of objects, thereby adopting adaptive handling strategies \cite{yuan2018active, baishya2016robust}. 
Despite this potential, many existing visuo-tactile sensors rely heavily on tracking embedded physical markers to estimate tangential forces and slip states \cite{dong2017improved, sferrazza2019design}. %
These markers inadvertently obscure fine surface textures, thereby limiting the sensor's ability to capture high-definition geometric features of the contact interface. %
Furthermore, existing slip datasets predominantly categorize contact into simplistic states, such as slip, non-slip, or pressing \cite{meier2016distinguishing}. %
Given that tactile slip during physical interaction is inherently a complex and dynamic process, such rudimentary classifications fall short of providing the nuanced state estimation required for advanced robotic manipulation. %

To address these limitations, we propose a novel visuo-tactile sensor specifically designed for marker-less, multi-task perception with high-precision texture sensitivity. %
The main contributions of this paper are summarized as follows:
\begin{table*}[t] 
    \centering
    \caption{COMPARISON OF EXISTING TACTILE SLIP DATASETS}
    \label{tab:dataset_comparison}
    \begin{tabularx}{\textwidth}{@{} l l c c X l @{}}
        \toprule
        \textbf{Reference} & \textbf{Sensor Type} & \textbf{Objects} & \textbf{Classes} & \textbf{Slip State Granularity} & \textbf{Dataset Scale} \\
        \midrule
        Meier \textit{et al.} \cite{meier2016distinguishing} 
        & Piezoresistive array 
        & -- 
        & 3 
        & Stable, Translational Slip, Rotation 
        & Time-series data \\

        Zhang \textit{et al.} \cite{zhang2018slip} 
        & GelSight (Marker-based) 
        & Multiple 
        & 2 
        & Slip, Non-slip (Binary) 
        & Video sequences \\

        James \textit{et al.} \cite{james2018slip} 
        & TacTip (Biomimetic) 
        & -- 
        & 2 
        & Slip, Non-slip (Binary) 
        & 40 sequences \\

        Zapata-Impata \textit{et al.} \cite{zapata2019tactile}
        & BioTac SP 
        & 11 
        & 7 
        & Contact, 4 Translation, 2 Rotation 
        & 300 frames \\

        STNet \cite{zhu2024stnet} 
        & Visuo-tactile 
        & -- 
        & 2 
        & Slip, Non-slip (Binary) 
        & Not specified \\

        \midrule
        \textbf{Ours} 
        & \textbf{Marker-less VTS} 
        & \textbf{15} 
        & \textbf{10} 
        & \textbf{Static, Pressing, 8-directional slip} 
        & \textbf{7925 sequences} \\
        \bottomrule
    \end{tabularx}
\end{table*}

\begin{itemize}
    \item \textbf{Novel Hardware Design:} We developed a marker-less visuo-tactile sensor that utilizes a customized internal RGB illumination scheme. %
    This design effectively avoids specular highlights and captures subtle micro-textures that are frequently obscured in marker-based systems, enabling highly accurate sub-millimeter depth reconstruction. %
    
    \item \textbf{A Comprehensive RGB-D Tactile Benchmark Dataset:} We constructed a multi-task dataset encompassing 15 distinct object categories for spatiotemporal slip and material classification. Crucially, we synchronously generated corresponding depth information for each data sample. This provides a dense depth map where every single pixel corresponds to a precise physical depth value, offering rich and aligned visuo-tactile representations. The complete dataset will be open-sourced upon publication. %
    
    \item \textbf{Unified Perception Framework:} We introduce a unified data-driven pipeline achieving simultaneous, real-time slip detection and tactile-based object class recognition. %
    Specifically, a dual-head TimeSformer network is designed for robust dynamic spatiotemporal slip detection (predicting both contact states and fine-grained slip directions), while a ResNet-50 backbone is utilized for high-precision static material classification. %
\end{itemize}

The remainder of this paper is organized as follows: Section II reviews the related work on data-driven slip detection and tactile-based object class recognition. Section III details the hardware design and the sensing principle for depth reconstruction. Section IV introduces the data collection process and the unified data-driven perception framework. Section V presents the experimental setups and analyzes the perception results. Finally, Section VI concludes the paper and discusses future work.

\begin{figure*}
    \centering
    \includegraphics[width=\textwidth]{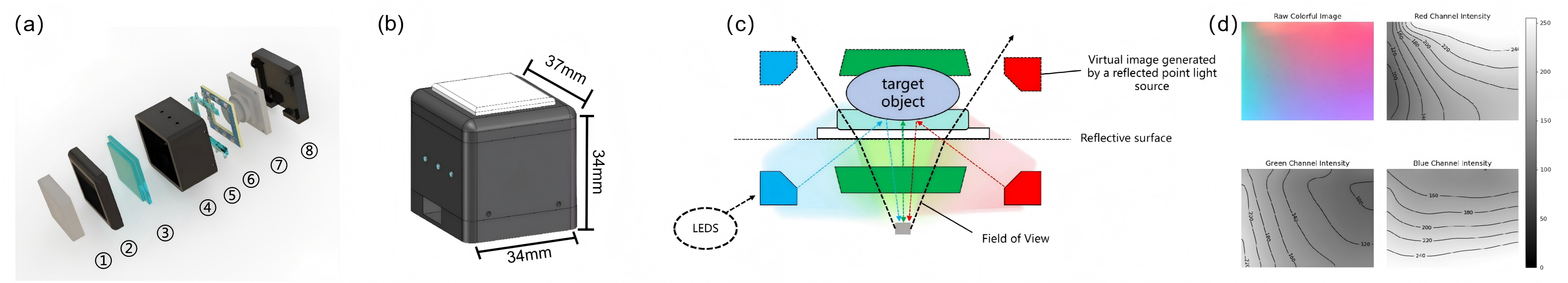}
    \caption{(a) Exploded view of the sensor components. \ding{172} Optically clear silicone coated with a white silicone pigment layer. \ding{173} Supports the elastomer. \ding{174} Transparent acrylic plate. \ding{175} Main housing. \ding{176} Translucent frosted diffuser. \ding{177} PCB for RGB illumination. \ding{178} Miniature camera. \ding{179} Base plate. (b) The overall dimensions of the sensor are 37 mm × 34 mm × 34 mm. (c) The internal optical path design, which effectively prevents specular highlights generated by the acrylic plate. (d) Actual image captured by the visuo-tactile sensor. Decomposing the image into individual RGB channels demonstrates a uniform intensity distribution and consistent brightness across all channels. }
\end{figure*}

\section{RELATED WORK}

\subsection{Data-Driven Slip Detection}
Slip detection is a critical technology to enable reactive grasping and stable object manipulation in robotics. With the advancement of deep learning, researchers have increasingly framed slip detection as a data-driven spatiotemporal perception problem. Early studies employed traditional CNN-LSTM architectures to fuse tactile and visual information, extracting slip events from consecutive image frames \cite{zhang2018slip}. Current mainstream approaches, however, integrate self-attention mechanisms and Vision Transformers to process visuo-tactile sequence data \cite{wang2023vitac}. For instance, recent works such as STNet utilize spatio-temporal fusion-based self-attention to precisely focus on core contact regions during dynamic interactions \cite{zhu2024stnet}. Furthermore, related studies indicate that simplistic binary slip classification is no longer sufficient for advanced manipulation tasks; detecting the specific slip direction is crucial for providing richer feedback to adjust grasping force. Unlike methods relying on shallow optical flow or traditional recurrent network frameworks, our work employs the TimeSformer architecture. Without the need for explicit marker tracking, this approach effectively captures fine-grained spatiotemporal textural shifts across 8 distinct slip directions. Concurrently, we constructed a comprehensive tactile dataset encompassing 15 distinct object categories. This dataset classifies contact into three primary states: slip, static, and pressing. Furthermore, to meet the stringent requirements for precise state estimation during dynamic manipulation, the ``slip" state is finely subdivided into eight directional categories at 45-degree intervals. The resulting slip detection dataset comprises 7925 sequences, with each sample containing 8 temporal frames (totaling 63,400 images).

\subsection{Tactile-based Object Class Recognition and Benchmark Datasets}
Identifying the physical properties of manipulated objects is another core application of tactile perception \cite{luo2017robotic}. Due to the rich spatial features inherent in their output data, visuo-tactile sensors exhibit superior performance potential in material classification tasks compared to traditional sensor arrays \cite{chortos2016pursuing,luo2018vitac}. Recent research has leveraged large-scale visuo-tactile datasets, such as the ``Touch in the Wild" framework \cite{chen2024touch}, demonstrating that massive, contact-rich datasets are indispensable for achieving fine-grained robotic manipulation. However, existing benchmark datasets for tactile-based object class recognition often suffer from limited scale and insufficient categorization, with some still relying on images obscured by physical markers \cite{sferrazza2019design}. To address the limitations of existing datasets in scale and categorization, we collected a dedicated tactile-based object class recognition dataset containing 19336 images to capture the unique visuo-tactile signatures of various surfaces, thereby validating the exceptional fidelity of our marker-less sensor.

\section{Hardware Design and Sensing Principle}

\subsection{Overall Design of the Visuo-Tactile Sensor} 
The visuo-tactile sensor proposed in this work features a highly compact and robust design, with overall dimensions of 37 mm × 34 mm × 34 mm, allowing for seamless integration into standard robotic grippers for closed-loop operation. Generally, the architecture of a visuo-tactile sensor comprises three fundamental modules: an elastomeric contact module, which serves as the physical interface for tactile perception; an illumination module, typically a printed circuit board (PCB) integrated with tri-color RGB LEDs; and a signal reception module, primarily a miniature camera capable of capturing RGB data.

As shown in Fig. 2(a), to facilitate contact tasks and the calibration of light intensity, the illumination and camera modules are integrated into a single base unit, while the elastomeric component is molded as an independent, detachable module. This modular design significantly enhances the sensor's suitability for long-term, continuous operations. When the elastomer sustains wear and tear from repeated frictional contact, replacing the contact module is all that is required to resume capturing tactile signals under identical lighting conditions.

Unlike conventional empirical illumination setups \cite{10870064}, the spatial arrangement of the light sources in our design has been meticulously engineered(as shown in Fig. 2(c)). The light sources are effectively isolated into distinct illumination zones on the PCB, and a light-diffusing filter is positioned above them to ensure a highly uniform distribution of light across the contact layer. Furthermore, because the light illuminates directly from beneath the elastomer, specular reflection from the acrylic support plate typically generates ghost images, which compromises the final imaging quality. To address this, we optimized the relative positioning of the contact module and the camera. This ensures that the camera's field of view completely covers the contact surface while keeping the ghost images entirely out of its focal range. Consequently, this design guarantees high illumination uniformity across the entire inner surface of the elastomer alongside excellent imaging quality.

We also quantified the intensity variation of each color channel by illuminating a standard flat surface. The resulting distribution, as shown in Fig. 2(d), aligns consistently with the spatial layout of the LEDs. This customized, uniform RGB illumination effectively suppresses specular highlights. It is specifically tailored for photometric stereo, thereby enabling the accurate reconstruction of 3D surface topography and the extraction of high-frequency micro-textures from markerless images.

\subsection{Sensor Fabrication}
The fabrication process begins with manufacturing the mechanical components and designing the electrical circuitry for the camera-based tactile sensor, as depicted in Fig. 2(a). Subsequently, the elastomer layer is cast, a reflective coating is applied, and the individual modules are assembled into a fully integrated sensor(as shown in Fig. 3).

To streamline the manufacturing process, all structural components and the casting molds for the elastomer were fabricated using a 3D printer. The elastomer casting process proceeded as follows: First, the upper housing of the elastic contact module and the acrylic plate were tightly fitted together and secured with adhesive. The 3D-printed silicone molds were then attached to the top and bottom of this assembled unit. A platinum-catalyzed, optically clear silicone (Smooth-On Solaris) was prepared with a 1:1 mixing ratio and poured into the mold. The setup was then vacuum-degassed to eliminate trapped air bubbles and left undisturbed at room temperature to cure.

\begin{figure*}
    \centering
    \includegraphics[width=1\linewidth]{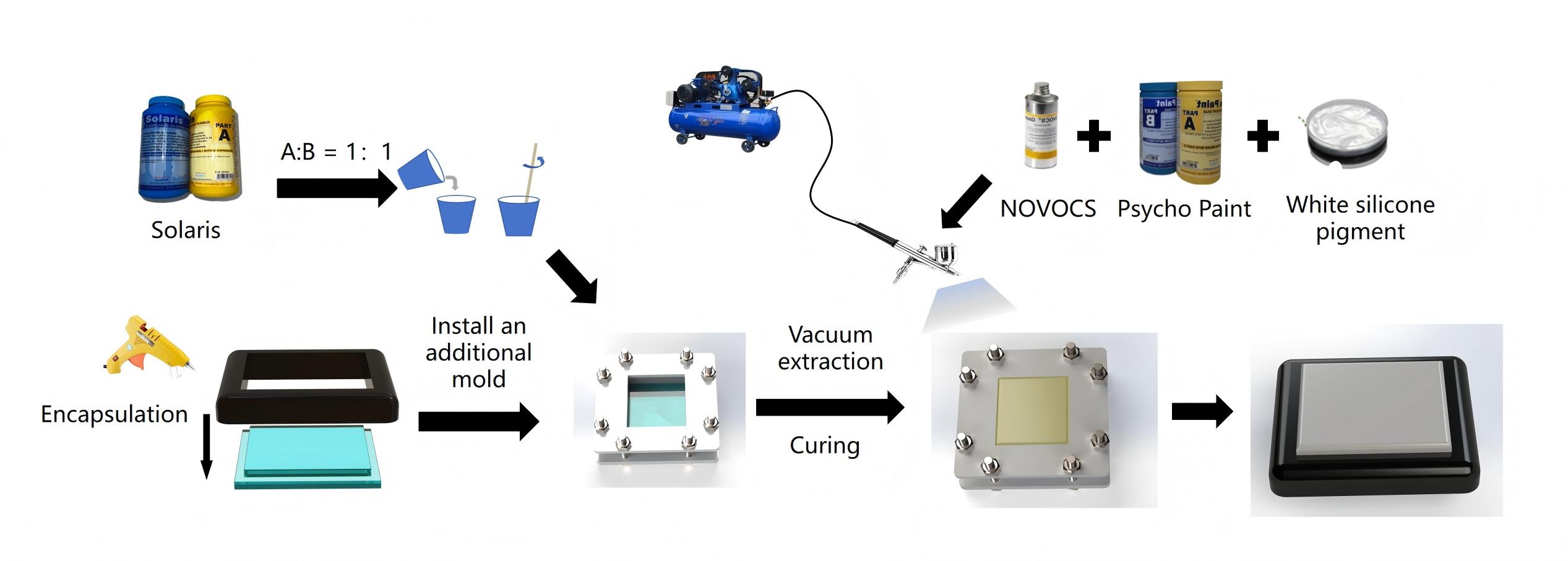}
    \caption{Fabrication process of the elastic contact module. The optically clear silicone (Smooth-On Solaris) was first mixed and poured into a customized 3D-printed mold, followed by curing. A custom matte reflection coating, blended from silicone base and white pigment, was then uniformly airbrushed onto the elastomer surface. The final assembled contact module features a compact and easily replaceable design.}
    \label{fig:placeholder}
\end{figure*}

Following the curing process, a custom coating mixture was prepared by blending white silicone pigment, a silicone paint base (Smooth-On Psycho Paint), and a silicone solvent (Smooth-On NOVOCS Thinner) in a volumetric ratio of 0.3:1:1:2. This mixture was evenly sprayed onto the elastomer surface to form a thin, matte white coating, which serves as the diffuse reflection layer. Finally, the components were demolded and readied for subsequent sensor integration.

\subsection{Depth Reconstruction}

To perform depth reconstruction \cite{zhang2025mlp}, we employ an empirical photometric stereo approach calibrated using a reference sphere. As shown in Fig. 4(b), the calibration process began by pressing a high-precision standard sphere with a known physical radius of $r = 3 \text{ mm}$ into the sensor's elastomer using a three-axis linear stage.

As shown by the geometric model in Fig. 4(a), let the contact center of the spherical impression in the image plane be denoted by the coordinates $(x_c, y_c)$. For any pixel $(x, y)$ within the contact region, its projected radial distance to the center is defined as $r = \sqrt{(x - x_c)^2 + (y - y_c)^2}$. The 3D depth profile $z(x, y)$ of the spherical impression follows the geometric equation:
\begin{equation}
    z(x, y) = \sqrt{R^2 - r^2}
\end{equation}

Since this geometric feature is completely known, we can precisely calculate the ground-truth surface gradients in the horizontal and vertical directions (denoted as $p_{ref}$ and $q_{ref}$, respectively) by taking the partial derivatives of $z(x, y)$ with respect to $x$ and $y$:
\begin{equation}
    p_{ref}(x, y) = \frac{\partial z}{\partial x} = \frac{-(x - x_c)}{\sqrt{R^2 - r^2}}
\end{equation}
\begin{equation}
    q_{ref}(x, y) = \frac{\partial z}{\partial y} = \frac{-(y - y_c)}{\sqrt{R^2 - r^2}}
\end{equation}

Corresponding to the projection features shown on the right side of Fig. 4(a), the camera captures the RGB intensity values $I(x, y) = [I_R, I_G, I_B]^T$ for each pixel within the impression area. To compensate for minor spatial non-uniformities in the illumination distribution, we explicitly incorporate the spatial coordinates $(x, y)$ of the pixels alongside their color intensities as inputs to train a continuous mapping function $f$:
\begin{equation}
    [p(x, y), q(x, y)] = f(I_R, I_G, I_B, x, y)
\end{equation}

During the calibration phase, this mapping relationship $f$ was established by fitting the input data to the ground-truth gradients $[p_{ref}, q_{ref}]$. During the actual tactile sensing of unknown objects, the captured RGB images are converted into dense gradient fields $p(x, y)$ and $q(x, y)$ using the calibrated mapping model $f$. Finally, by integrating these gradient fields using a Poisson solver, the continuous, high-fidelity depth map $z(x, y)$ of the target object is reconstructed:
\begin{equation}
    \nabla^2 z(x, y) = \frac{\partial p}{\partial x} + \frac{\partial q}{\partial y}
\end{equation}

\begin{figure}
    \centering
    \includegraphics[width=1\linewidth]{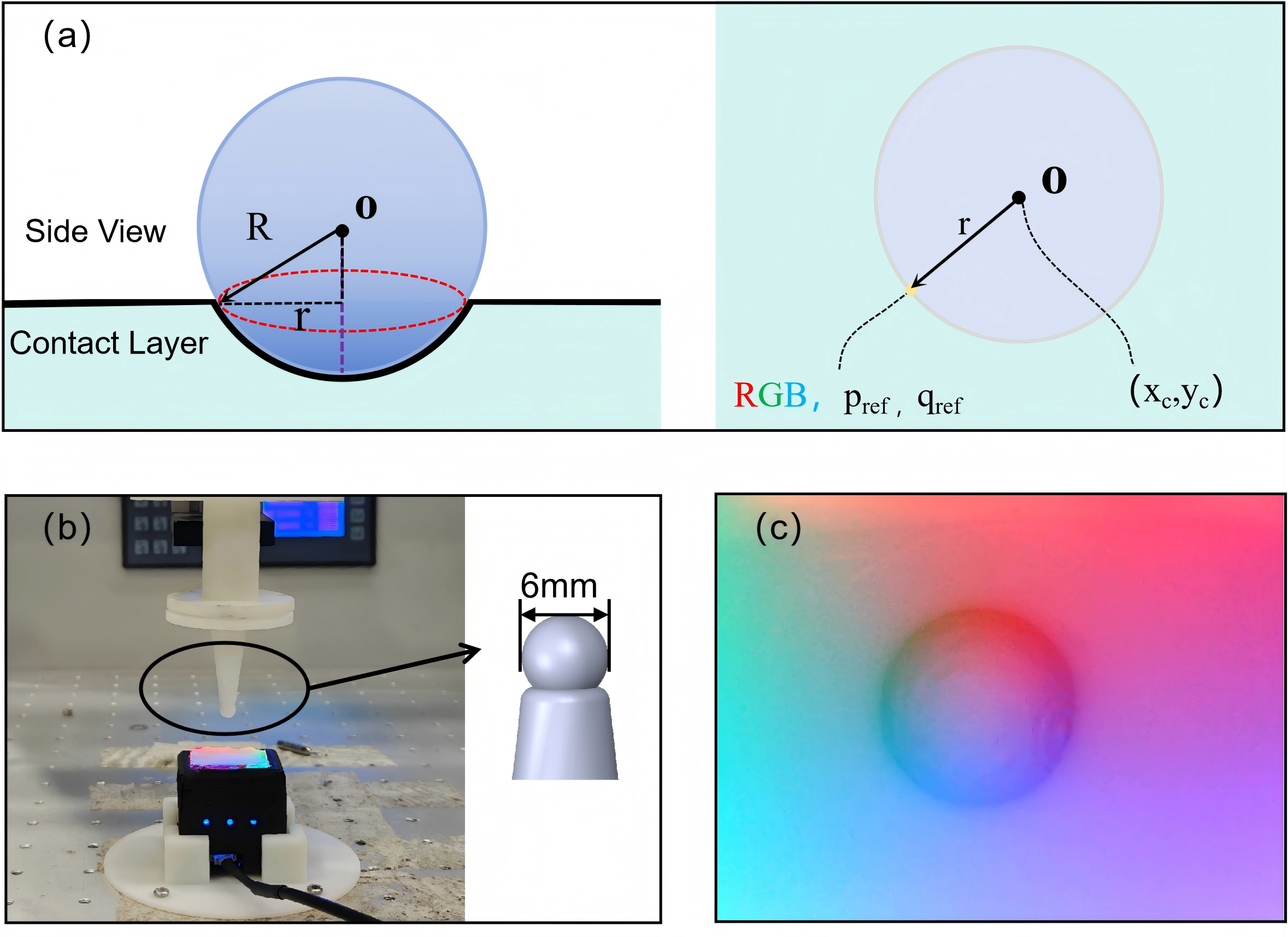}
    \caption{Empirical photometric stereo calibration using a reference sphere. (a) Geometric model of the spherical indentation. The side view illustrates the physical radius $R$ and the projected radial distance $r$, while the top view demonstrates the mapping relationship from the contact center $(x_c, y_c)$ and RGB intensities to the surface gradients. (b) The physical calibration setup, where a standard reference sphere with a diameter of 6 mm is pressed into the sensor using a precision three-axis linear stage. (c) The corresponding RGB image captured by the internal miniature camera during the spherical indentation.}
    \label{fig:3D}
\end{figure}

\section{DATA COLLECTION AND DATASET}

To evaluate the performance of the proposed visuo-tactile sensor in micro-texture extraction and dynamic physical interactions, we constructed a comprehensive tactile benchmark dataset comprising a static material classification subset and a dynamic spatiotemporal slip subset. Furthermore, by leveraging the sensor's highly accurate depth reconstruction, we synchronously converted all raw visuo-tactile samples across both subsets into one-to-one corresponding dense depth maps, yielding a comprehensive RGB-D tactile dataset. Crucially, the customized uniform illumination acts as a physical prior for the downstream Spatiotemporal Transformer. By eliminating specular highlights, it ensures strict spatial and temporal consistency of the captured micro-textures, preventing the self-attention mechanism from overfitting to lighting noise.

\subsection{Material and Object Classification Dataset}

\begin{figure*}
    \centering
    \includegraphics[width=1\linewidth]{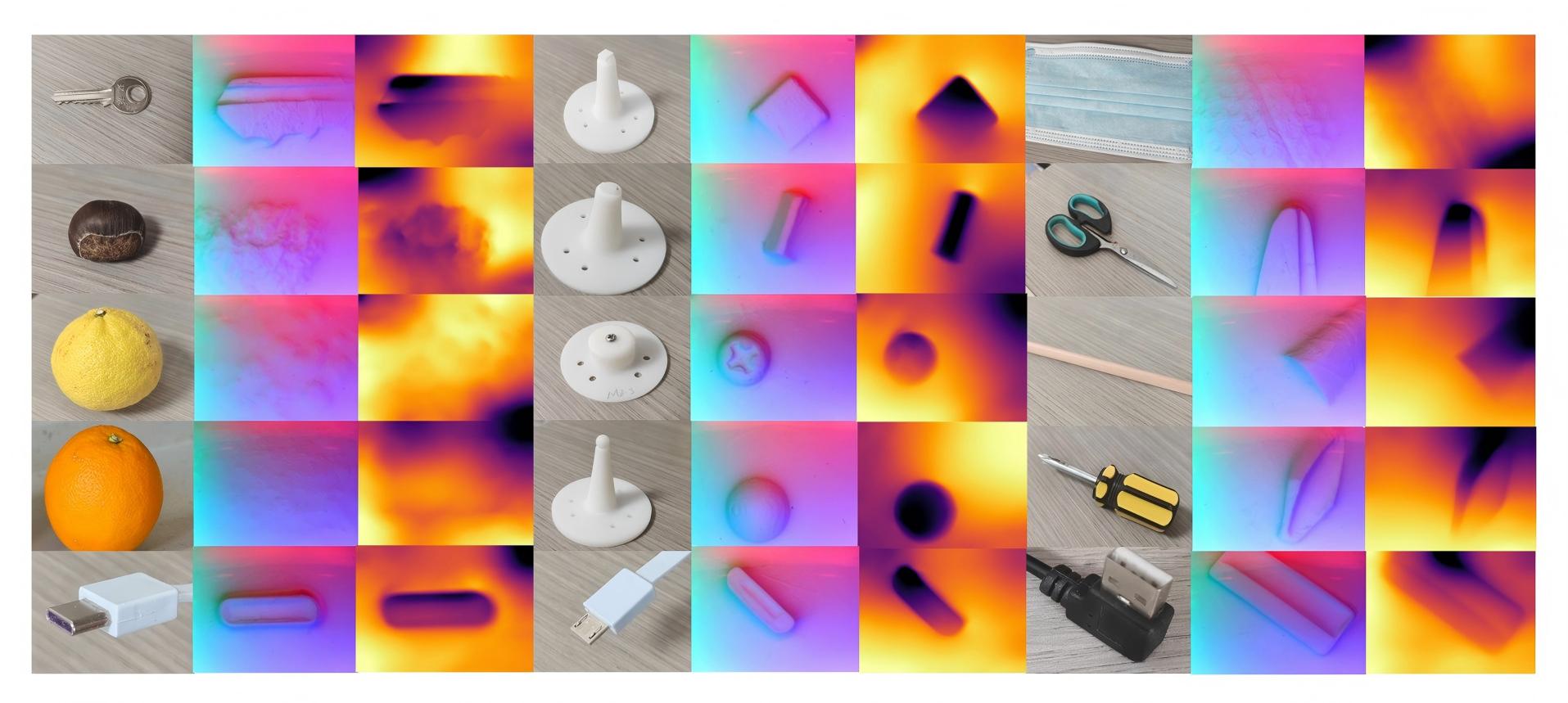} 
    \caption{The 15 representative daily objects, along with their corresponding high-resolution tactile images and reconstructed depth maps. Benefiting from the uniform illumination, micro-textures and geometric contours are presented with exceptional clarity without specular highlight interference.}
    \label{fig:objects_dataset}
\end{figure*}

To rigorously evaluate the sensor's perception limits regarding complex topographies, we selected 15 representative daily objects (as shown in Fig. \ref{fig:objects_dataset}). These samples encompass standard geometric indenters, hardware tools, precision electronic connectors, and natural organic textures, comprehensively testing the sensor's spatial resolution.

During collection, samples were captured across diverse contact locations, spatial poses, and pressing depths, yielding approximately 19336 high-resolution tactile images. Benefiting from the uniform RGB illumination, high-frequency details are stably captured without specular interference. This provides an exceptionally high-quality data foundation for learning robust material features.

\subsection{Spatiotemporal Slip Dataset}

To capture complex evolutionary patterns during dynamic interactions, we constructed a large-scale omnidirectional slip dataset using the aforementioned 15 objects. This dataset consists of 7925 high-fidelity slip sequences (totaling around 63,400 images). To balance computational efficiency and temporal feature integrity, each sample is organized as an 8-frame continuous image sequence.

Regarding state categorization, the dataset is divided into three core physical states: Pressing, Static, and Slip. To meet advanced directional perception requirements, the ``Slip" state is further refined into 8 distinct directional categories at $45^\circ$ intervals. By integrating slip samples across diverse objects, the dataset forces the model to ignore macro-geometric identities and focus exclusively on high-frequency spatiotemporal texture shifts. This provides a rigorous benchmark for verifying the robustness of the Spatiotemporal Transformer.

\begin{figure}
    \centering
    \includegraphics[width=1\linewidth]{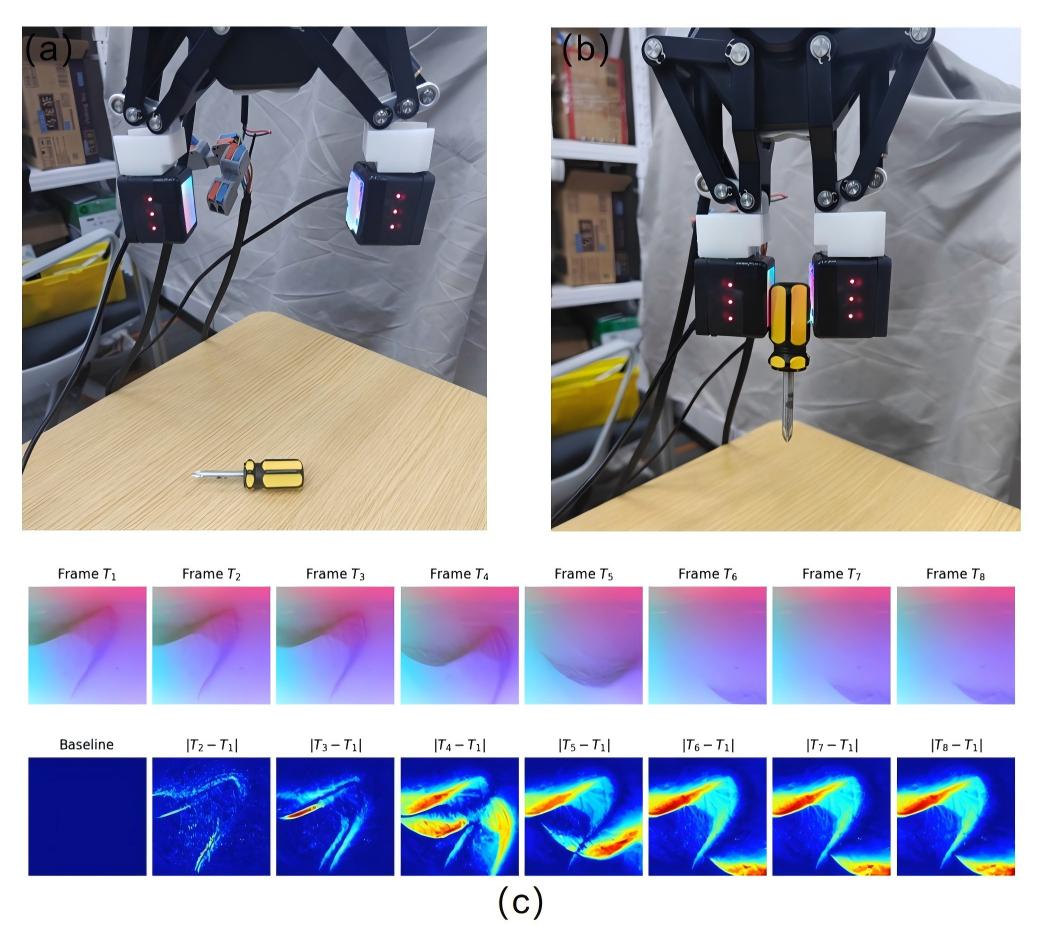} 
    \caption{Visualization of the real-world grasping process and the corresponding tactile data. The figure displays a sequence of continuous raw RGB tactile images captured during dynamic interaction, alongside the corresponding temporal difference maps between consecutive frames.}
    \label{fig:slip_sequence}
\end{figure}

\subsection{Data-Driven Perception Framework}

we propose a unified data-driven perception framework. This architecture is tailored for both dynamic spatiotemporal slip tracking and static material classification.

\begin{figure*}
    \centering
    \includegraphics[width=\textwidth]{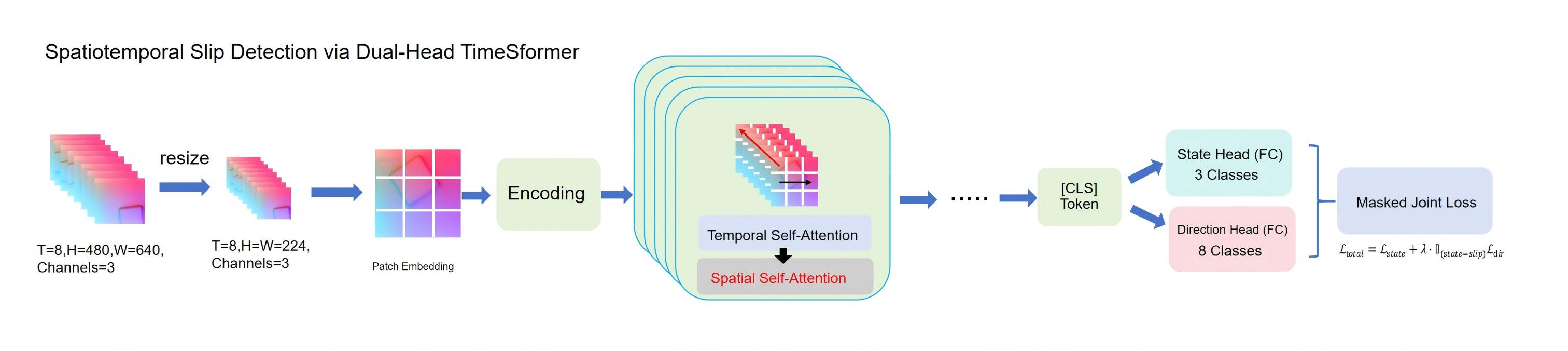} 
    \caption{The overall architecture of the proposed spatiotemporal slip detection network based on a dual-head TimeSformer. The input sequence is processed through patch embedding and divided space-time attention blocks to extract dynamic features. A global token is then utilized by dual FC heads to predict the contact state and slip direction, optimized by a masked joint loss.}
    \label{fig:timesformer_arch}
\end{figure*}

\textbf{1) Spatiotemporal Slip Detection via Dual-Head TimeSformer:} 
Tactile slip is intrinsically the spatial displacement of micro-textures at the contact interface over a brief period. To achieve an optimal trade-off between the global receptive field and computational efficiency, we introduce a marker-less slip detection pipeline based on a dual-head TimeSformer (as shown in Fig. \ref{fig:timesformer_arch}).

Given a raw 8-frame tactile sequence tensor with the shape of $T = 8, H = 480, W = 640, C = 3$, the input is first resized to $224 \times 224$. The sequence is then divided into non-overlapping spatial patches, linearly projected, and augmented with both spatial and temporal positional encodings. The core of this network is the stacked \textit{Divided Space-Time Attention} blocks, which strictly alternate between computing temporal self-attention (across frames) and spatial self-attention (within the same frame). Benefiting from the consistent micro-textures captured under uniform illumination, the temporal attention mechanism can sharply track physical displacements without being distracted by lighting artifacts.

After passing through multiple attention blocks, the global token is extracted and fed into a dual-head fully connected (FC) architecture to predict the contact state and direction simultaneously. The \textit{State Head} outputs probabilities for 3 primary classes (Static, Press, Slip), while the \textit{Direction Head} predicts the slip vector across 8 directional classes. To optimize this multi-task learning process, we propose a masked joint loss function:
\begin{equation}
    \mathcal{L}_{total} = \mathcal{L}_{state} + \lambda \cdot \mathbb{I}_{(\text{state}=\text{slip})} \mathcal{L}_{dir}
\end{equation}
where $\lambda$ is a balancing weight, and the indicator function $\mathbb{I}$ strictly ensures that the direction loss $\mathcal{L}_{dir}$ is only propagated when the ground-truth state is ``Slip". This masking strategy effectively prevents noise interference from static or purely pressing states.

\textbf{2) Static Material Classification via ResNet-50:}
Unlike dynamic slip, material classification relies on static spatial geometry and micro-texture features at the exact moment of contact. By leveraging the strict spatial invariance provided by the uniform illumination, we employ a ResNet-50 architecture pretrained on ImageNet as the backbone for this task. 

The network takes a single tactile image (resized to $224 \times 224$) as input. During training, we apply color jittering and geometric augmentations to further enhance the model's robustness against minor optical fluctuations. The fully connected layer is modified to output the 15 material categories. Additionally, to provide interpretability for the network's decisions, we integrate Gradient-weighted Class Activation Mapping (Grad-CAM) to visualize the specific high-frequency micro-features (sharp boundaries or organic pores) that contribute most to the network's predictions, thereby rigorously validating the physical fidelity of our sensor hardware.

\section{EXPERIMENTS AND RESULTS}

To evaluate the proposed uniformly-illuminated visuo-tactile sensor, we conducted extensive experiments focusing on 3D depth reconstruction, material classification, and spatiotemporal slip detection.

\subsection{Experimental Setup}
For the 3D depth reconstruction task, the ground-truth depth was analytically derived during the calibration process using a standard sphere of a known radius, based on the previously discussed empirical photometric stereo formulation. Conversely, the predicted 3D morphology was reconstructed by first mapping the RGB images to obtain surface gradients, and subsequently solving the Poisson equation over these output gradients.
Regarding the experimental platform, a RealMan RM75-6F robotic arm was utilized, equipped with a DH-Robotics AG-160-95 gripper integrating the aforementioned visuo-tactile sensors. All deep learning models were trained and evaluated on a workstation equipped with a single NVIDIA RTX 5090 GPU. The software environment was configured with Python 3.10.19 and PyTorch 2.7.1 (CUDA 12.8). Furthermore, for the spatiotemporal slip detection task, our model was initialized with the pre-trained facebook/timesformer-base-finetuned-ssv2 weights to fully leverage its robust spatiotemporal feature representation capabilities.

\subsection{Depth Reconstruction}

\begin{figure}[htbp]
    \centering
    \includegraphics[width=1\linewidth]{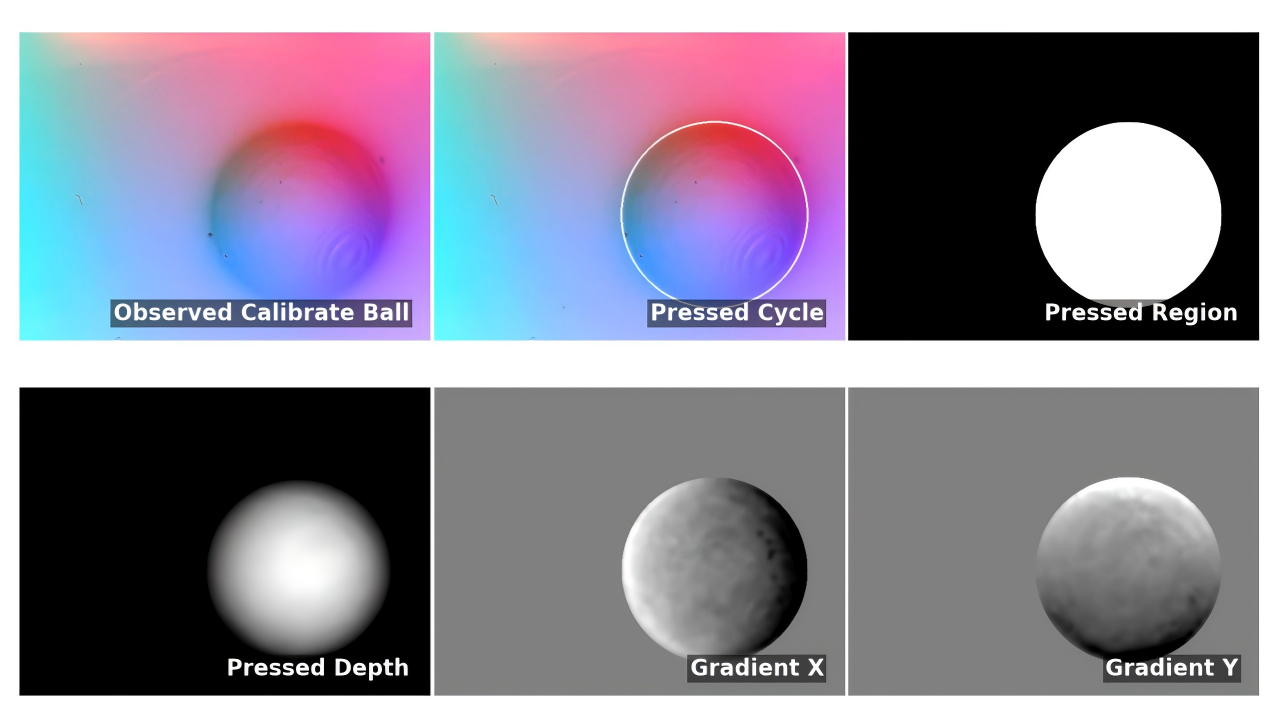}
    \caption{The visuo-tactile perception pipeline for 3D depth reconstruction. The sub-millimeter precision (RMSE = 0.072 mm) relies on the high-quality gradient decoupling and the masked Poisson integration.}
    \label{fig:depth_recon}
\end{figure}

Accurate depth reconstruction is fundamental to visuo-tactile perception \cite{zhang2025mlp, s22176470}. Fig. \ref{fig:depth_recon} illustrates our reconstruction pipeline using a standard spherical indenter. From the raw RGB observation, a binary contact mask is extracted. The neural network then predicts the high-frequency planar gradients ($X$ and $Y$ axes), followed by a masked Poisson integration. This method enforces a strict zero-potential baseline for non-contact regions, yielding a pure depth map.

To isolate true topological accuracy from potential mechanical misalignments, we evaluated the {Plane-aligned Root Mean Square Error (RMSE)}. Our sensor achieved an exceptional mean Plane-aligned RMSE of {0.072 mm} ($\sim 72 \mu m$). This sub-millimeter precision confirms that our customized uniform illumination effectively eliminates specular high lights providing a highly reliable geometric baseline for downstream perception tasks.

\subsection{Dynamic Spatiotemporal Slip Detection}

To evaluate dynamic perception capabilities, we trained a dual-head TimeSformer architecture on our sequential tactile dataset. Using raw RGB sequences, the model achieved exceptional baseline accuracies of 99.69\% and 98.86\% for 3-class contact state prediction and 8-class slip direction classification, respectively. Furthermore, when evaluated on dynamic depth sequences reconstructed from the RGB data, the model achieved accuracies of 94.51\% and 70.57\% for the 3-class state prediction and 8-class direction classification tasks, respectively.

\begin{figure}
    \centering
    \includegraphics[width=1\linewidth]{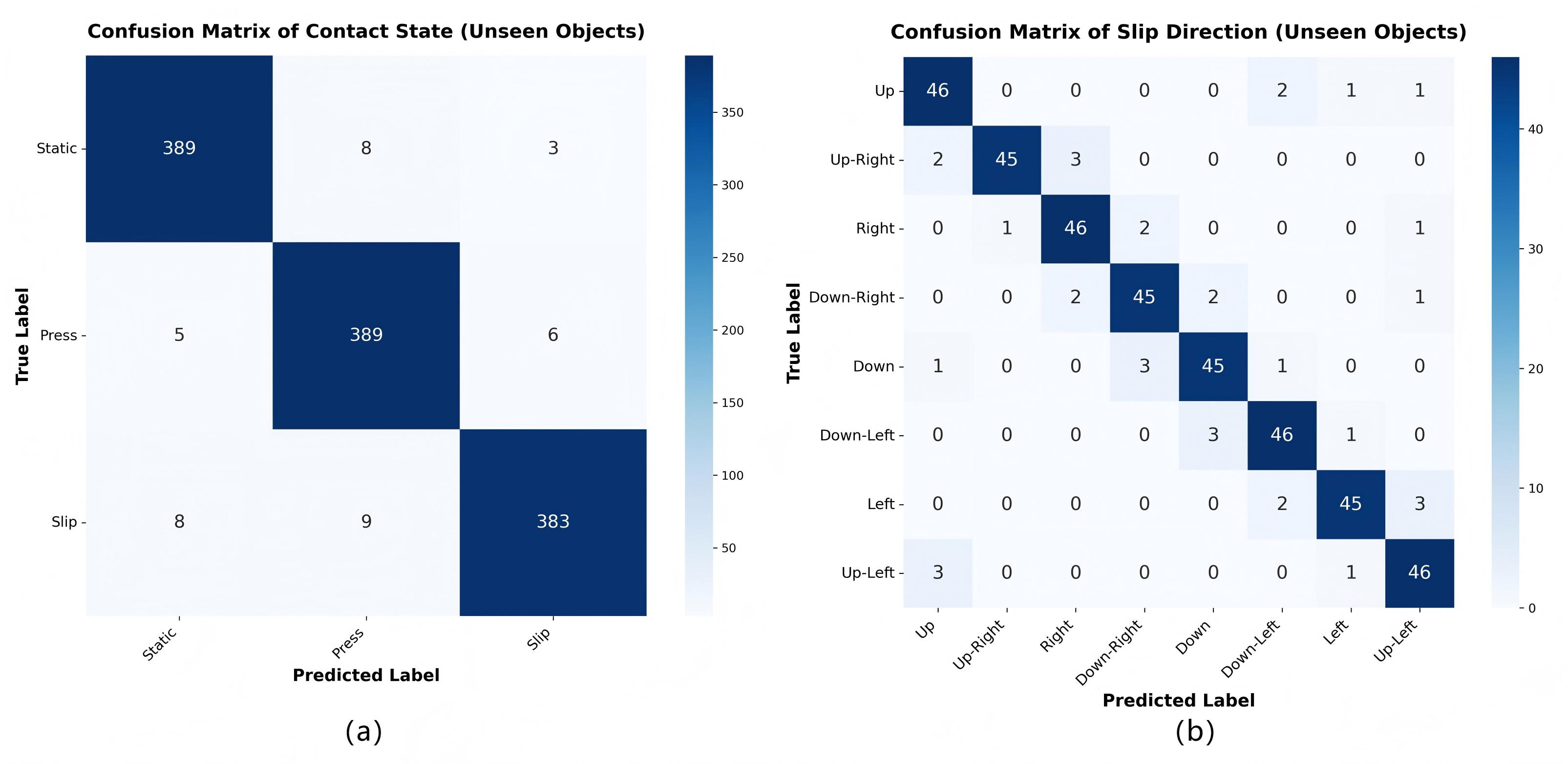}
    \caption{Confusion matrices for dynamic spatiotemporal slip detection on unseen objects. (a) Results for the 3-class contact state prediction (Static, Press, and Slip). (b) Results for the fine-grained 8-class slip direction classification.}
    \label{fig:tactile}
\end{figure}

To rigorously assess the generalization capabilities of the highly accurate model trained on RGB sequences to unknown physical properties, we evaluated 5 unseen objects using 1,200 independent sequences (400 static, 400 pressing, and 400 slipping). As shown in Fig. \ref{fig:tactile}(a), the model achieved a 95.5\% accuracy in contact state classification. Subsequently, the 400 sequences predicted as ``slip'' were passed to the direction head, maintaining a robust accuracy of 91.5\% across the 8 specific directions (as shown in Fig. \ref{fig:tactile}(b)). The physically intuitive error distribution (predominantly localized in adjacent angular directions) confirms that our uniform illumination perfectly preserves the high-frequency micro-textures essential for precise displacement tracking.

\begin{figure}[htbp]
    \centering
    \includegraphics[width=\linewidth]{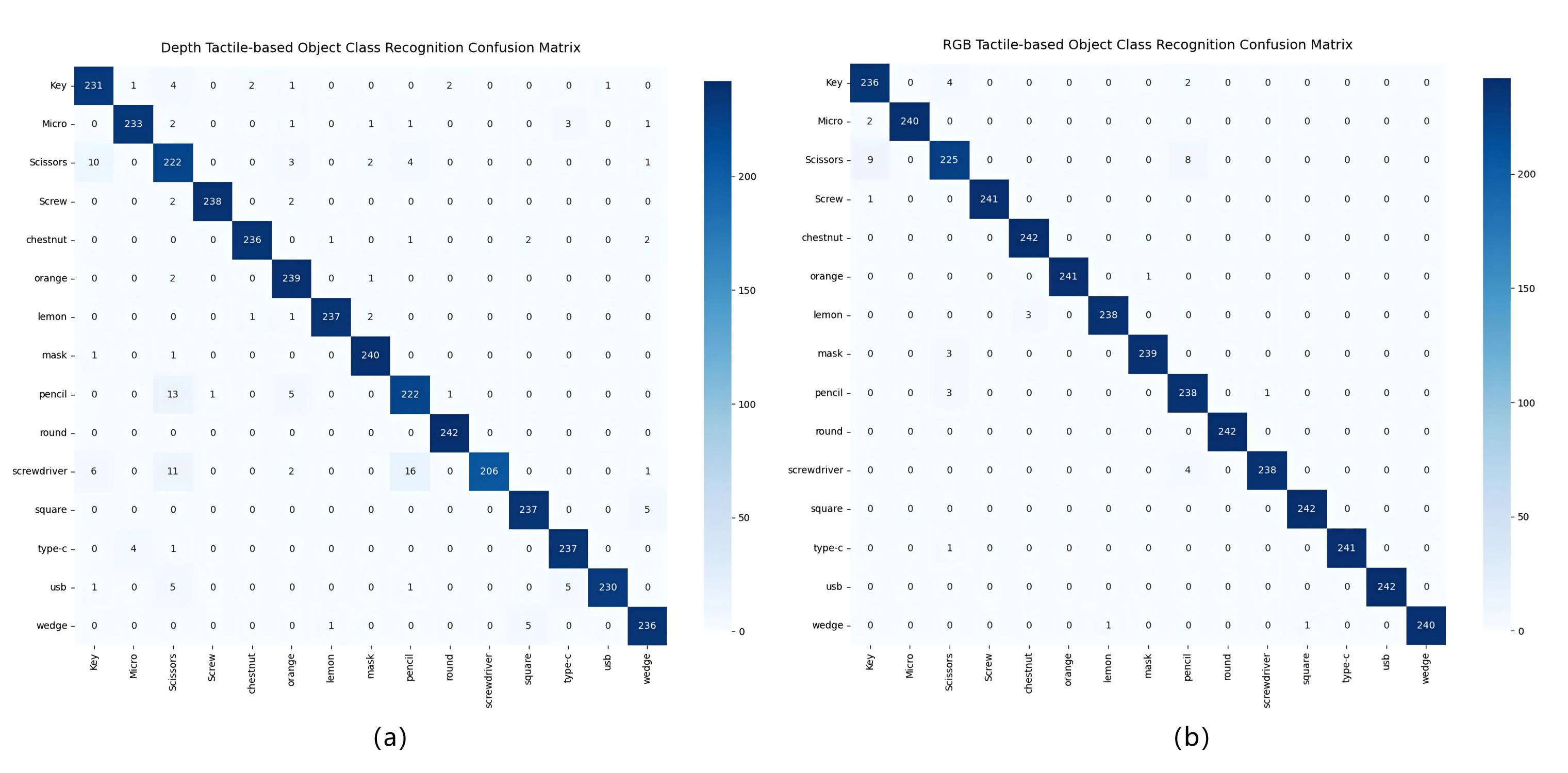}
    \caption{Confusion matrices for dynamic and static visuo-tactile perception tasks, respectively showing the 3-class contact state prediction on unseen objects, the 8-class slip direction classification on unseen objects, and the 15-class static  tactile-based object class recognition.}
    \label{fig:tactile_recog_cm}
\end{figure}

\subsection{Static Tactile-based Object Class Recognition}

Extracting fine-grained spatial representations is fundamental for precise object manipulation. In this study, we rigorously evaluated the sensor's static perception limits by classifying 15 categories, encompassing a diverse range of daily objects, textured fruits, and standard geometric shapes (as shown in Fig. \ref{fig:objects_dataset}). 

During the experiment, single-frame raw RGB tactile images and their corresponding reconstructed depth maps were independently fed into a ResNet-50 backbone. Benefiting from the physical prior provided by the uniform RGB illumination, the model effectively extracted pure textural and 3D topological features, free from specular highlight noise. 

As shown in the confusion matrices, the architecture successfully learned highly discriminative multi-modal representations. Across approximately 4,600 test samples, the RGB modality achieved a superior accuracy of {98.8\%} due to its rich micro-textural details, while the depth modality reached an accuracy of {96.06\%}, validating the high fidelity of our macroscopic geometric reconstruction. This dual-modal performance provides a robust data foundation for subsequent complex manipulation tasks.

\begin{table}[htbp]
\centering
\caption{Summary of Visuo-Tactile Perception Accuracies}
\label{tab:accuracy_summary}
\begin{tabular}{llc}
\hline
\textbf{Perception Task} & \textbf{Test Condition (Modality)} & \textbf{Accuracy} \\ \hline
\multirow{2}{*}{Static Recognition (15-class)} & RGB (Baseline) & \textbf{98.80\%} \\
 & Depth-only & 96.06\% \\ \hline
\multirow{3}{*}{Contact State (3-class)} & RGB (Baseline) & \textbf{99.68\%} \\
 & RGB (Unseen Objects) & \textbf{95.50\%} \\
 & Depth-only & 94.51\% \\ \hline
\multirow{3}{*}{Slip Direction (8-class)} & RGB (Baseline) & \textbf{96.95\%} \\
 & RGB (Unseen Objects) & \textbf{91.50\%}  \\
 & Depth-only & 71.53\% \\ \hline
\end{tabular}
\end{table}

\section{CONCLUSIONS}

This paper presents a novel uniformly illuminated visuo-tactile sensor featuring high-precision texture sensitivity, alongside a unified perception framework for robust slip detection and material classification. Through an optimized optical design, our sensor overcomes the severe artifacts caused by specular reflections inherent in traditional vision-based depth estimation, successfully achieving high-fidelity 3D topological reconstruction with an exceptionally low Root Mean Square Error (RMSE) of {0.072 mm}. 

To support this framework, we constructed a large-scale tactile dataset comprising 7925 high-fidelity slip sequences (totaling 63,400 images) and 19,336 static images across 15 diverse objects. For dynamic perception, the introduced dual-head TimeSformer architecture achieved exceptional baseline accuracies of {99.68\%} and {96.95\%} for contact state prediction and slip direction classification, respectively. Crucially, the model demonstrated robust generalization on unseen objects, maintaining high accuracies of {95.5\%} for state classification and {91.5\%} for fine-grained direction detection. Furthermore, in static recognition tasks, the model achieved an outstanding {98.8\%} accuracy across 15 material categories, validating its superior spatial representation capabilities.

In future work, we plan to integrate this sensor into a multi-fingered dexterous robotic hand. By fusing the high-frequency tactile feedback from our framework with global vision and proprioceptive data, we aim to enable real-time, closed-loop manipulation and grasp adaptation in highly challenging dynamic environments.






\bibliographystyle{IEEEtran}

\bibliography{REFERENCES}

@article{dahiya2010tactile,
  title={Tactile sensing—from sensors to systems},
  author={Dahiya, Ravinder S and Metta, Giorgio and Valle, Maurizio and Sandini, Giulio},
  journal={IEEE Transactions on Robotics},
  volume={26},
  number={1},
  pages={1--20},
  year={2010},
  publisher={IEEE}
}

@article{li2020review,
  title={A review of tactile information: Perception and action through touch},
  author={Li, Qiang and Kroemer, Oliver and Su, Zhe and Veiga, Filipe F and others},
  journal={IEEE Transactions on Robotics},
  volume={36},
  number={6},
  pages={1619--1634},
  year={2020},
  publisher={IEEE}
}

@article{luo2017robotic,
  title={Robotic tactile perception of object properties: A review},
  author={Luo, Shan and Bimbo, Joao and Dahiya, Ravinder and Liu, Hongbin},
  journal={Mechatronics},
  volume={48},
  pages={54--67},
  year={2017},
  publisher={Elsevier}
}

@article{zhang2022hardware,
  title={Hardware technology of vision-based tactile sensor: A review},
  author={Zhang, Shixin and Chen, Zixi and Gao, Yuan and Wan, Weiwei and Shan, Jianhua and Xue, Hongxiang and Sun, Fuchun and Yang, Yiyong and Fang, Bin},
  journal={IEEE Sensors Journal},
  volume={22},
  number={22},
  pages={21132--21147},
  year={2022},
  publisher={IEEE}
}

@article{calandra2018more,
  title={More than a feeling: Learning to grasp and regrasp using vision and touch},
  author={Calandra, Roberto and Owens, Andrew and Upadhyaya, Manu and Yuan, Wenzhen and Lin, Justin and Adelson, Edward H and Levine, Sergey},
  journal={IEEE Robotics and Automation Letters},
  volume={3},
  number={4},
  pages={3300--3307},
  year={2018},
  publisher={IEEE}
}

@article{yuan2017gelsight,
  title={GelSight: High-resolution tactile sensors for estimating geometry and force},
  author={Yuan, Wenzhen and Dong, Siyuan and Adelson, Edward H},
  journal={Sensors},
  volume={17},
  number={12},
  pages={2762},
  year={2017},
  publisher={MDPI}
}

@article{ward2018tactip,
  title={The TacTip family: Soft optical tactile sensors with 3D-printed biomimetic morphologies},
  author={Ward-Cherrier, Benjamin and Pestell, Nicolas and Cramphorn, Luke and Winstone, Benjamin and Giannaccini, Maria E and Burgess, Jonathan and Lepora, Nathan F},
  journal={Soft robotics},
  volume={5},
  number={2},
  pages={204--215},
  year={2018},
  publisher={Mary Ann Liebert, Inc.}
}

@article{chortos2016pursuing,
  title={Pursuing prosthetic electronic skin},
  author={Chortos, Alex and Liu, Jian and Bao, Zhenan},
  journal={Nature materials},
  volume={15},
  number={9},
  pages={937--950},
  year={2016},
  publisher={Nature Publishing Group}
}

@article{lepora2021soft,
  title={Soft biomimetic optical tactile sensing with the TacTip: A review},
  author={Lepora, Nathan F},
  journal={IEEE Sensors Journal},
  volume={21},
  number={19},
  pages={21131--21143},
  year={2021},
  publisher={IEEE}
}

@article{james2018slip,
  title={Slip detection with a biomimetic tactile sensor},
  author={James, J Ward and Pestell, Nicolas and Lepora, Nathan F},
  journal={IEEE Robotics and Automation Letters},
  volume={3},
  number={4},
  pages={3340--3346},
  year={2018},
  publisher={IEEE}
}

@article{zapata2019tactile,
  title={Tactile spatio-temporal neural networks for object slipping detection},
  author={Zapata-Impata, Brayan S and Gil, Pablo and Torres, Fernando},
  journal={Sensors},
  volume={19},
  number={23},
  pages={5218},
  year={2019},
  publisher={MDPI}
}

@inproceedings{yuan2018active,
  title={Active clothing material perception using tactile sensing and deep learning},
  author={Yuan, Wenzhen and Mo, Yuchen and Wang, Shaoxiong and Adelson, Edward H},
  booktitle={2018 IEEE International Conference on Robotics and Automation (ICRA)},
  pages={4842--4849},
  year={2018},
  organization={IEEE}
}

@inproceedings{baishya2016robust,
  title={Robust material classification with a tactile soft sensor},
  author={Baishya, S and B{\"a}cher, M},
  booktitle={2016 IEEE/RSJ International Conference on Intelligent Robots and Systems (IROS)},
  pages={8--15},
  year={2016},
  organization={IEEE}
}

@inproceedings{dong2017improved,
  title={Improved GelSight tactile sensor for measuring geometry and slip},
  author={Dong, Siyuan and Yuan, Wenzhen and Adelson, Edward H},
  booktitle={2017 IEEE/RSJ International Conference on Intelligent Robots and Systems (IROS)},
  pages={137--144},
  year={2017},
  organization={IEEE}
}

@article{sferrazza2019design,
  title={Design, motivation and evaluation of a full-resolution optical tactile sensor},
  author={Sferrazza, Carmelo and D’Andrea, Raffaello},
  journal={Sensors},
  volume={19},
  number={4},
  pages={928},
  year={2019},
  publisher={MDPI}
}

@inproceedings{meier2016distinguishing,
  title={Distinguishing sliding from slipping during object manipulation},
  author={Meier, Martin and Walck, Gilwoo and Haschke, Robert and Ritter, Helge},
  booktitle={2016 IEEE/RSJ International Conference on Intelligent Robots and Systems (IROS)},
  pages={5257--5262},
  year={2016},
  organization={IEEE}
}

@article{zhu2024stnet,
  title={STNet: Spatio-Temporal Fusion-Based Self-Attention for Slip Detection in Visuo-Tactile Sensors},
  author={Zhu, Yi and others}, 
  journal={IEEE Sensors Journal},
  volume={24},
  number={8},
  pages={13588--13598},
  year={2024},
  publisher={IEEE}
}

@inproceedings{chen2024touch,
  title={Touch in the Wild: Learning Fine-Grained Manipulation with a Portable Visuo-Tactile Gripper},
  author={Chen, Xinyu and others},
  booktitle={8th Annual Conference on Robot Learning (CoRL)},
  year={2024}
}

@article{wang2023vitac,
  title={ViTac: Vision transformer for tactile sensing},
  author={Wang, Z. and others},
  journal={IEEE Robotics and Automation Letters},
  volume={8},
  number={5},
  pages={2571--2578},
  year={2023},
  publisher={IEEE}
}

@article{Li_2025,
   title={Classification of Vision-Based Tactile Sensors: A Review},
   volume={25},
   ISSN={2379-9153},
   DOI={10.1109/jsen.2025.3599236},
   number={19},
   journal={IEEE Sensors Journal},
   publisher={Institute of Electrical and Electronics Engineers (IEEE)},
   author={Li, Haoran and Lin, Yijiong and Lu, Chenghua and Yang, Max and Psomopoulou, Efi and Lepora, Nathan F.},
   year={2025},
   month=oct, pages={35672–35686} }

@Article{s22176470,
AUTHOR = {Ji, Jingjing and Liu, Yuting and Ma, Huan},
TITLE = {Model-Based 3D Contact Geometry Perception for Visual Tactile Sensor},
JOURNAL = {Sensors},
VOLUME = {22},
YEAR = {2022},
NUMBER = {17},
ARTICLE-NUMBER = {6470},
PubMedID = {36080929},
ISSN = {1424-8220},
DOI = {10.3390/s22176470}
}

@ARTICLE{10870064,
  author={Lu, Chuang and Liang, Ziting and Stoyanov, Danail and Stilli, Agostino},
  journal={IEEE Sensors Journal}, 
  title={GelPoLight: A Novel Visual-Tactile Sensor Based on Photometric Stereo With Point Lighting}, 
  year={2025},
  volume={25},
  number={6},
  pages={9575-9584},
  doi={10.1109/JSEN.2025.3535080}}

@InProceedings{zhang2025mlp,
author="Zhang, Shixin and Sun, Yuhao and Sun, Funchun and Liu, Huaping and Yang, Yiyong and Fang, Bin",
title="MLP-Depth: An Improved Visuo-Tactile 3D Reconstruction Method Applied to TIRgel Sensor",
booktitle="Intelligent Robotics and Applications",
year="2025",
publisher="Springer Nature Singapore",
pages="275--285",
isbn="978-981-96-0795-2"
}

@inproceedings{chen2023plasticine,
  title={Plasticine Manipulation Simulation with Optical Tactile Sensing},
  author={Chen, Zixi and Zhang, Shixin and Sun, Yuhao and Luo, Shan and Sun, Fuchun and Fang, Bin},
  booktitle={ICRA ViTac Workshop},
  year={2023}
}

@article{zhang2025artificialskin,
author = {Zhang, Shixin and Yang, Yiyong and Sun, Yuhao and Liu, Nailong and Sun, Fuchun and Fang, Bin},
title = {Artificial Skin Based on Visuo-Tactile Sensing for 3D Shape Reconstruction: Material, Method, and Evaluation},
journal = {Advanced Functional Materials},
volume = {35},
number = {1},
pages = {2411686},
year = {2025}
}

@article{luo2018vitac,
author = {Luo, Shan and Yuan, Wenzhen and Adelson, Edward and Cohn, Anthony and Fuentes, Raul},
journal = {arXiv preprint arXiv:1802.07490},
year = {2018},
month = {02},
pages = {},
title = {ViTac: Feature Sharing Between Vision and Tactile Sensing for Cloth Texture Recognition},
doi = {10.48550/arXiv.1802.07490}
}

@article{sun2025tactile,
  title={Tactile data generation and applications based on visuo-tactile sensors: A review},
  author={Sun, Yuhao and Cheng, Ning and Zhang, Shixin and Li, Wenzhuang and Yang, Lingyue and Cui, Shaowei and Liu, Huaping and Sun, Fuchun and Zhang, Jianwei and Guo, Di and others},
  journal={Information Fusion},
  volume={121},
  pages={103162},
  year={2025},
  publisher={Elsevier}
}

@inproceedings{zhang2018slip,
  title={Slip detection with combined tactile and visual information},
  author={Zhang, Hao and others},
  booktitle={2018 IEEE International Conference on Robotics and Automation (ICRA)},
  pages={1--6},
  year={2018},
  organization={IEEE}
}

\end{document}